\documentclass{article} 
\usepackage{iclr2026_conference,times}

\usepackage{amsmath,amsfonts,bm}

\def\eqref#1{equation~\ref{#1}}

\def\1{\bm{1}}

\DeclareMathAlphabet{\mathsfit}{\encodingdefault}{\sfdefault}{m}{sl}
\SetMathAlphabet{\mathsfit}{bold}{\encodingdefault}{\sfdefault}{bx}{n}

\usepackage{hyperref}
\usepackage{url}
\usepackage{amsmath}
\usepackage{amssymb}
\usepackage{amsfonts}
\usepackage{mathtools}
\usepackage{bm}
\usepackage{booktabs}
\usepackage{multirow}
\usepackage[table]{xcolor}

\definecolor{lightgray}{gray}{0.92}

\usepackage{algorithm}
\usepackage{algpseudocode}

\title{Hierarchy-Guided Multimodal Representation Learning for Taxonomic Inference}

\author{
Sk Miraj Ahmed$^{1}$, Xi Yu$^{1}$, Yunqi Li$^{2}$\thanks{This author contributed to this work while at Brookhaven National Laboratory.}, Yuewei Lin$^{1}$, Wei Xu$^{1}$ \\
$^{1}$Computing and Data Sciences, Brookhaven National Laboratory, Upton, NY 11973, USA \\
$^{2}$Rutgers University, New Brunswick, NJ, USA \\
$^{1}$\texttt{\{sahmed3, xyu1, ywlin, xuw\}@bnl.gov} \\
$^{2}$\texttt{yunqi.li@rutgers.edu}
}

\iclrfinalcopy 
\begin{document}

\maketitle

\begin{abstract}
Accurate biodiversity identification from large-scale field data is a foundational problem with direct impact on ecology, conservation, and environmental monitoring. In practice, the core task is \emph{taxonomic prediction}---inferring order, family, genus, or species from imperfect inputs such as specimen images, DNA barcodes, or both. Existing multimodal methods often treat taxonomy as a flat label space and therefore fail to encode the hierarchical structure of biological classification, which is critical for robustness under noise and missing modalities. We present two end-to-end variants for hierarchy-aware multimodal learning: \textbf{CLiBD-HiR}, which introduces \textit{Hierarchical Information Regularization (HiR)} to shape embedding geometry across taxonomic levels, yielding structured and noise-robust representations; and \textbf{CLiBD-HiR-Fuse}, which additionally trains a \textit{lightweight fusion predictor} that supports image-only, DNA-only, or joint inference and is resilient to modality corruption. Across large-scale biodiversity benchmarks, our approach improves taxonomic classification accuracy by over 14\% compared to strong multimodal baselines, with particularly large gains under partial and corrupted DNA conditions. These results highlight that explicitly encoding biological hierarchy, together with flexible fusion, is key for practical biodiversity foundation models.
\end{abstract}

\section{Introduction}

Biodiversity research \cite{van2018inaturalist} increasingly relies on large-scale multimodal data, including specimen images, DNA barcodes, and auxiliary taxonomic metadata. While curated datasets enable controlled benchmarking, real-world deployments are far less controlled. In particular, DNA barcodes obtained through large-scale sequencing pipelines (e.g., BOLD \cite{ratnasingham2007bold}) can exhibit partial reads, ambiguous bases, and sequencing artifacts \cite{medigue1999detecting, meiklejohn2019assessment}, and field-collected specimen images are often degraded by cluttered backgrounds, occlusions, lighting variation, motion blur \cite{nguyen2024sawit}, or low signal-to-noise acquisition. Bridging this gap between curated evaluation and imperfect operational inputs makes robust taxonomic inference from heterogeneous and noisy modalities a central and unresolved challenge in applied biodiversity science.

A key recent step toward unifying these modalities is \textbf{CLIBD}~\cite{gong2024clibd}, which to our knowledge is the only prior work that explicitly aligns \emph{images, DNA barcodes, and taxonomic text} in a shared embedding space at \emph{very large scale}, and is therefore our primary state-of-the-art reference. In practice, however, cross-modal retrieval typically surfaces candidate matches that still require downstream verification by human experts, whereas operational pipelines ultimately benefit from reliable \emph{taxonomic prediction} from one or more available modalities. More importantly, existing multimodal objectives commonly treat taxonomy as a flat label space and rely on standard contrastive learning without explicitly encoding biological hierarchy. As a result, learned embeddings may lack hierarchy-consistent geometry: closely related taxa are not guaranteed to be nearby, and perturbations from noise or missing data can lead to unpredictable errors across taxonomic levels. This issue is particularly acute when one modality---most commonly DNA---is partially corrupted or unavailable, a routine scenario in large-scale biodiversity repositories. Finally, most existing approaches do not explicitly model \emph{adaptive image--DNA fusion}, even though DNA alone may be insufficient in practice and complementary morphological cues from images can be critical for resolving fine-grained taxa \cite{cong2017coi}.

\begin{figure*}[t]
    \centering
    \includegraphics[width=\textwidth]{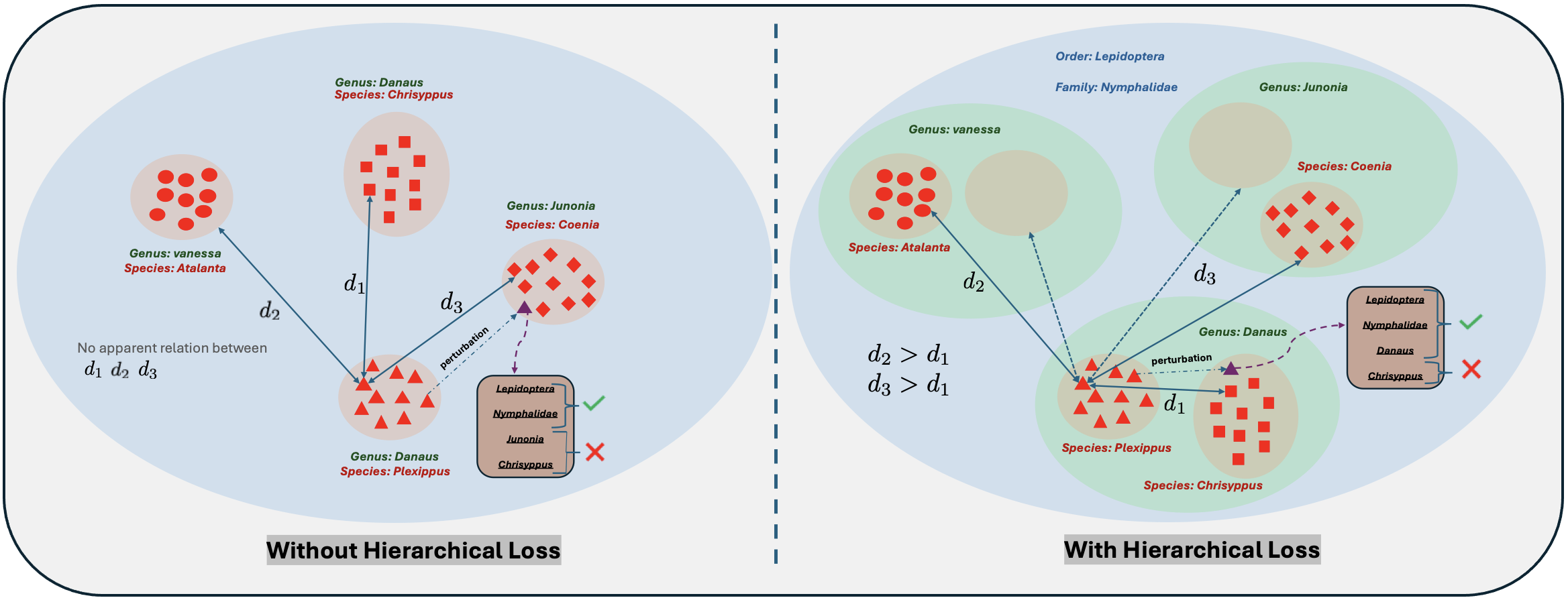}
    \caption{\textbf{Effect of hierarchical regularization on embedding geometry and noise robustness (CLIBD-HiR, variant~1).}
\textbf{Left:} without the HiR loss, standard contrastive training treats mismatched taxa uniformly, yielding no explicit geometric relationship between intra-genus distances ($d_1$; different species within the same genus) and inter-genus / higher-level distances ($d_2,d_3$). Under realistic noise, a perturbed query embedding may drift across arbitrary clusters, leading to errors that can propagate to higher taxonomic ranks. 
\textbf{Right:} with HiR, the loss explicitly enforces a hierarchy-consistent structure ($d_1 < d_2 < d_3$), so nearby neighborhoods reflect taxonomic proximity. Consequently, even when noise causes a species-level mistake, predictions are more likely to remain correct at coarser levels (genus/family/order), improving robustness.}
    \label{fig:main_framework_1}
\end{figure*}

In this work, we build on CLIBD~\cite{gong2024clibd} and present a taxonomy-aware multimodal framework for robust taxonomic prediction from images, DNA barcodes, and their combination under realistic noise. Our approach is instantiated as two complementary end-to-end variants of CLIBD (Algo 1 and 2) that (i) inject taxonomic hierarchy into representation learning via a hierarchy-aware objective, and (ii) optionally train an explicit image--DNA fusion predictor for joint inference.

\textbf{Algo 1: CLIBD-HiR (structured, noise-robust representation learning).}
Our first variant targets a core failure mode of prior contrastive methods: the lack of hierarchy-consistent structure in the learned representation space. We propose \textbf{Hierarchical Information Regularization (HiR)}, which injects taxonomic hierarchy directly into representation learning and explicitly \emph{shapes the embedding geometry} during training. Samples sharing coarser taxa (e.g., family or genus) are encouraged to remain close, while finer distinctions (e.g., species) are learned without collapsing higher-level neighborhoods. This hierarchical organization acts as an intrinsic noise stabilizer: when a noisy sample drifts away from its species cluster due to image corruption or barcode degradation, HiR still anchors it to the correct higher-level neighborhood, limiting catastrophic semantic drift (Fig.~\ref{fig:main_framework_1}).

\begin{figure*}[t]
    \centering
    \includegraphics[width=\textwidth]{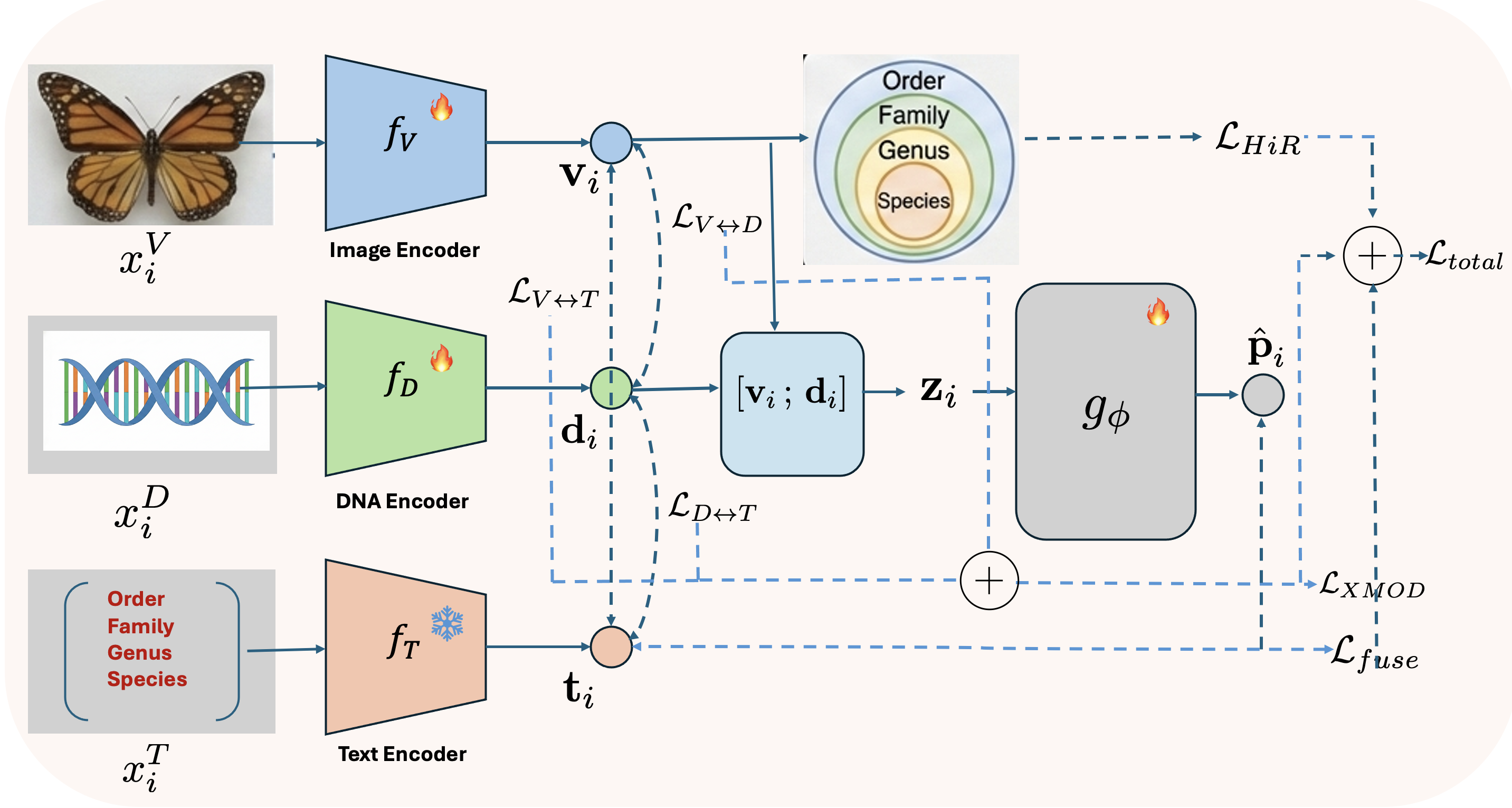}
    \caption{\textbf{CLIBD-HiR-Fuse framework (Algorithm variant 2).} 
Given a specimen image and its DNA barcode, we encode each modality with an image encoder and a DNA encoder, and embed the taxonomy prompt with a frozen BioCLIP text encoder. 
We align image--text and DNA--text representations using CLIP-style contrastive learning, and enforce hierarchy-aware structure with a hierarchical loss over augmented image views. 
A lightweight GatedFusion module adaptively combines image and DNA embeddings into a fused representation, which is additionally aligned to the fixed text embedding space via a fused-to-text contrastive objective.}

    \label{fig:main_framework_2}
\end{figure*}

\textbf{Algo 2: CLIBD-HiR-Fuse (adaptive fusion for variable modality quality).}
Our second variant extends CLIBD-HiR with a lightweight fusion predictor trained jointly with the encoders. This is motivated by deployment realities where available evidence varies by sample---some specimens have only images, some only barcodes, and some both, often with differing degrees of corruption. CLIBD-HiR-Fuse (Fig. ~\ref{fig:main_framework_2}) supports image-only, DNA-only, and fused image+DNA inference, and leverages the hierarchy-aware aligned space to better combine complementary signals when one modality is unreliable.

Together, hierarchy-guided regularization (Algo~1) and adaptive fusion (Algo~2) produce noise-resilient multimodal models that better match real-world biodiversity workflows. Across large-scale benchmarks, our approach improves taxonomic prediction accuracy over CLIBD and strong fusion baselines, with particularly large gains under DNA corruption and low-quality imaging.

\textbf{Our contributions are threefold:}

\noindent$\bullet$ We introduce \textbf{Hierarchical Information Regularization (HiR)}, a taxonomy-aware objective that explicitly shapes embedding geometry and improves robustness to noisy and partially corrupted inputs.\\
\noindent$\bullet$ We present two end-to-end variants: \textbf{CLIBD-HiR} (Algo~1), a structured embedding learner optimized for hierarchical taxonomic prediction, and \textbf{CLIBD-HiR-Fuse} (Algo~2), which adds an adaptive fusion predictor supporting image-only, DNA-only, and image+DNA inference under varying modality quality.\\
\noindent$\bullet$ We demonstrate consistent improvements in taxonomic prediction across biodiversity benchmarks, with especially large gains in noise-dominated regimes.

\section{Related Work}
\label{sec:related_work}

\paragraph{Foundation models for biodiversity and multimodal alignment.}
Recent progress in foundation models has enabled transferable representations via large-scale pretraining and multimodal alignment. CLIP-style training aligns vision and language to support zero-shot transfer \citep{radford2021clip,cherti2023openclip_scaling}, and newer multimodal alignment frameworks extend this idea beyond image--text to jointly embed heterogeneous modalities. For example, ImageBind learns a shared space across multiple sensory streams (e.g., image, text, audio, depth, IMU) using paired data \citep{girdhar2023imagebind}, and related efforts such as ``X-CLIP''/``LanguageBind''-style models extend CLIP-like alignment to additional modalities. In biodiversity, BioCLIP adapts CLIP-style pretraining toward organism-centric visual recognition \citep{stevens2024bioclip}, while CLIBD is a key step toward \emph{multimodal} biodiversity foundation modeling by aligning specimen images, DNA barcodes, and taxonomic text in a shared embedding space \citep{gong2024clibd,bioscan1m}. These backbones provide strong representations, but are typically optimized for retrieval-style alignment and do not explicitly enforce hierarchy-consistent geometry or robustness to modality degradation.

\paragraph{Taxonomy-aware representation learning.}
Taxonomic labels are inherently hierarchical (order--family--genus--species), motivating objectives that respect coarse-to-fine structure beyond flat classification. Prior work has explored hierarchy-aware losses and hierarchical contrastive learning to impose semantic structure directly in the embedding space \citep{khosla2020supcon,zhang2022usealllabels}. Our HiR regularization builds on this line by injecting taxonomic hierarchy into multimodal alignment, shaping neighborhoods so that nearby embeddings reflect biological relatedness and improving robustness when fine-grained cues are noisy or incomplete.

\paragraph{Multimodal fusion under noisy modalities.}
Beyond alignment, practical biodiversity applications often benefit from combining complementary evidence across modalities. Recent fusion strategies include uncertainty-aware fusion that explicitly improves robustness to noisy unimodal representations \citep{Gao_2024_CVPR}, as well as dynamic routing / mixture-of-experts style fusion with learned gating that adapts fusion behavior to the input \citep{Cao_2023_ICCV,Han_2024_FuseMoE}. We build on these ideas with a lightweight gated fusion head that adaptively mixes image and DNA embeddings, and evaluate it against naive averaging under clean and degraded modality conditions.

\paragraph{Robust modeling of barcodes and field imagery.}
DNA barcodes encountered in operational pipelines can be imperfect (e.g., substitutions/indels, ambiguous bases, partial reads), motivating noise-aware preprocessing and modeling \cite{medigue1999detecting}; similarly, field imagery is affected by background clutter, occlusion, illumination/pose changes, and motion blur, which can degrade fine-grained recognition \cite{nguyen2024sawit}. Our work is complementary: rather than relying on noise-specific training data, we stabilize multimodal prediction by enforcing hierarchy-consistent geometry and enabling adaptive fusion to leverage complementary evidence when one modality is degraded.

\section{Method}
\label{sec:method}

We consider triplets $(x_i^{V}, x_i^{D}, x_i^{T})$ consisting of an image, a DNA barcode, and a textual taxonomy description for specimen $i$. Our goal is twofold: (1) learn a shared multimodal embedding space where visual, DNA, and text representations are aligned in a hierarchy-aware, noise-robust manner; and (2) learn a fusion network that directly predicts taxonomy from jointly using image and DNA features.

We denote the encoders by
\[
\mathbf{v}_i = f_V(x_i^{V}), \quad
\mathbf{d}_i = f_D(x_i^{D}), \quad
\mathbf{t}_i = f_T(x_i^{T}),
\]
where $\mathbf{v}_i, \mathbf{d}_i, \mathbf{t}_i \in \mathbb{R}^d$ are $\ell_2$-normalized embeddings.
Each specimen is also annotated with hierarchical taxonomic labels
\[
\mathbf{y}_i = \big( y_i^{(1)}, y_i^{(2)}, \dots, y_i^{(L)} \big),
\]
e.g., order, family, genus, species for $L=4$.

Our framework is trained in two stages:
(1) multimodal pretraining with symmetric cross-modal contrastive losses plus an image-only Hierarchical Information Regularization (HiR) loss; and
(2) post-hoc training of an MLP fusion head that predicts taxonomy from visual and DNA embeddings.

\subsection{Cross-Modal Contrastive Objectives}
Following CLIBD~\cite{gong2024clibd}, we align modalities with symmetric InfoNCE. For a batch of size $N$, define $s(\mathbf{a}_i,\mathbf{b}_j)=\mathbf{a}_i^\top\mathbf{b}_j/\tau$. The directed loss is
\begin{equation}
\mathcal{L}_{A\rightarrow B}=-\frac{1}{N}\sum_{i=1}^{N}\log\frac{\exp(s(\mathbf{a}_i,\mathbf{b}_i))}{\sum_{j=1}^{N}\exp(s(\mathbf{a}_i,\mathbf{b}_j))},
\end{equation}
and $\mathcal{L}_{A\leftrightarrow B}=\mathcal{L}_{A\rightarrow B}+\mathcal{L}_{B\rightarrow A}$. We use pairs $(V,T)$, $(V,D)$, and $(D,T)$, and combine them as
\begin{equation}
\mathcal{L}_{\text{XMOD}}=\lambda_{VT}\mathcal{L}_{V\leftrightarrow T}+\lambda_{VD}\mathcal{L}_{V\leftrightarrow D}+\lambda_{DT}\mathcal{L}_{D\leftrightarrow T}.
\end{equation}

\subsection{Hierarchical Information Regularization for Images}

While the cross-modal objective encourages alignment between modalities, it does not explicitly encode the taxonomy hierarchy, nor does it guarantee robustness when one modality is noisy or partially corrupted. To address this, we introduce an image-only hierarchical contrastive loss inspired by the HiConE objective from \cite{zhang2022usealllabels}.

For each taxonomic level $\ell \in \{1,\dots,L\}$ (e.g., order, family, genus, species), we treat images sharing the same label $y_i^{(\ell)}$ as positives at level $\ell$. Let $\mathcal{P}_i^{(\ell)}$ be the set of indices $j \neq i$ such that $y_j^{(\ell)} = y_i^{(\ell)}$, and let $\mathcal{N}_i^{(\ell)}$ be the remaining images in the batch.
For an anchor $i$ and a positive $j \in \mathcal{P}_i^{(\ell)}$, we define the \emph{pair-wise} supervised contrastive loss at level $\ell$ as
\begin{equation}
\ell^{(\ell)}(i,j)
=
- \log
\frac{\exp\big( s(\mathbf{v}_i, \mathbf{v}_j) \big)}
{\sum_{k \in \mathcal{P}_i^{(\ell)} \cup \mathcal{N}_i^{(\ell)}}
\exp\big( s(\mathbf{v}_i, \mathbf{v}_k) \big)}.
\label{eq:pair_loss_level}
\end{equation}
The standard level-$\ell$ supervised contrastive loss is then
\begin{equation}
\mathcal{L}_{\text{SupCon}}^{(\ell)}
= \frac{1}{N} \sum_{i=1}^{N}
\frac{1}{|\mathcal{P}_i^{(\ell)}|}
\sum_{j \in \mathcal{P}_i^{(\ell)}} \ell^{(\ell)}(i,j).
\end{equation}

Following HiConE, we further enforce that \emph{finer} taxonomic levels cannot be optimized in a way that violates coarser-level structure. To this end, we compute, for each level $\ell$, the maximum pair-wise loss over all positive pairs at that level:
\begin{equation}
m^{(\ell)}
=
\max_{i} \;
\max_{j \in \mathcal{P}_i^{(\ell)}} \;
\ell^{(\ell)}(i,j),
\label{eq:max_level_loss}
\end{equation}
and define a \emph{hierarchically rectified} pair loss
\begin{equation}
\tilde{\ell}^{(1)}(i,j) = \ell^{(1)}(i,j), \quad
\tilde{\ell}^{(\ell)}(i,j)
=
\max\big( \ell^{(\ell)}(i,j), \; m^{(\ell-1)} \big)
\;\;\text{for } \ell > 1.
\label{eq:rectified_pair_loss}
\end{equation}
Intuitively, this loss prevents the model from minimizing fine-grained losses while coarser-level structure is still poorly organized. Concretely, if the loss of a fine-level positive pair (e.g., same species) becomes \emph{smaller} than the largest loss among positives at the immediately coarser level (e.g., same genus), we \emph{clamp} the fine-level loss to that coarser-level maximum. This ensures that fine-level positives are not allowed to be optimized ``ahead'' of the coarser level: the model must first reduce the worst-case within-genus (or within-family) positive loss before further tightening species-level clusters. Our final Hierarchical Information Regularization (HiR) loss aggregates these rectified pair losses across levels:
\begin{equation}
\mathcal{L}_{\text{HiR}}
=
\sum_{\ell=1}^{L} \alpha_\ell
\left[
\frac{1}{N} \sum_{i=1}^{N}
\frac{1}{|\mathcal{P}_i^{(\ell)}|}
\sum_{j \in \mathcal{P}_i^{(\ell)}}
\tilde{\ell}^{(\ell)}(i,j)
\right],
\label{eq:hir_final}
\end{equation}
where $\alpha_\ell$ are non-negative weights (we use uniform weights in our experiments).

This hierarchical structure makes the visual encoder \emph{noise-robust}: if an embedding is perturbed such that it drifts away from its species cluster, the loss still anchors it using genus and family supervision, and the max-rectification in Eq.~\ref{eq:rectified_pair_loss} prevents the optimizer from overfitting to noisy fine-grained labels while ignoring coarser, more reliable signals. As a result, even under noisy supervision or mismatched modalities, the image representation preserves higher-level semantic consistency (Fig.~\ref{fig:main_framework_1}).






\subsection{Objectives: Two End-to-End Variants (Algo 1 vs. Algo 2)}

We instantiate our framework in two variants that share the same multimodal alignment and hierarchical regularization backbone, but differ in whether an explicit fusion predictor is trained.

\paragraph{Algo 1: CLiBD-HiR (structured, noise-robust representation learning).}
In Algo 1, we train the encoders end-to-end using the sum of cross-modal contrastive alignment and hierarchical image regularization:
\begin{equation}
\mathcal{L}_{\text{total}}^{(1)}
= \mathcal{L}_{\text{XMOD}}
+ \lambda_{\text{HiR}} \mathcal{L}_{\text{HiR}},
\label{eq:total_algo1}
\end{equation}
where $\lambda_{\text{HiR}}$ balances hierarchical image regularization against multimodal alignment.
This objective encourages a hierarchy-consistent embedding geometry, improving robustness when one modality is noisy or partially corrupted, and yielding strong taxonomy classification via nearest-neighbor or linear probing in the learned space. (Full pseudocde in Algorithm~\ref{alg:clibd_hir_e2e}).

\paragraph{Algo 2: CLiBD-HiR-Fuse (flexible fusion with missing-modality support).}
Algo 2 augments Algo 1 with a lightweight fusion module trained \emph{jointly} with the encoders under a single objective:
\begin{equation}
\mathcal{L}_{\text{total}}^{(2)}
= \mathcal{L}_{\text{XMOD}}
+ \lambda_{\text{HiR}} \mathcal{L}_{\text{HiR}}
+ \lambda_{\text{fuse}} \mathcal{L}_{\text{fuse}},
\label{eq:total_algo2}
\end{equation}
where $\lambda_{\text{fuse}}$ controls the contribution of fusion loss $\mathcal{L}_{\text{fuse}}$ (Eq.~\ref{eq:fuse_loss}).
This variant provides a direct prediction interface that is applicable when only one modality is available (image-only or DNA-only) and when both modalities are available, while remaining noise-resilient due to the shared aligned and hierarchy-aware representation. (Full pseudocde in Algorithm~\ref{alg:clibd_hir_fuse_e2e})

\subsection{Flexible Multimodal Fusion for Taxonomy Prediction (Algo 2)}

Algo 2 introduces a lightweight fusion predictor $g_\phi$ that maps available non-text modality embeddings to taxonomy logits.
Given image and DNA embeddings, we concatenate them as $\mathbf{z}_i=[\mathbf{v}_i;\mathbf{d}_i]\in\mathbb{R}^{2d}$ and define the fused representation as
\begin{equation}
\hat{\mathbf{p}}_i=
\mathbf{v}_i \ \text{(image-only)},\quad
\mathbf{d}_i \ \text{(DNA-only)},\quad
g_\phi(\mathbf{z}_i) \ \text{(image+DNA)}.
\label{eq:fuse_modes}
\end{equation}
We train $g_\phi$ with a supervised cross-entropy loss at level $\ell^\star$,
\begin{equation}
\mathcal{L}_{\text{fuse}}
= - \frac{1}{N} \sum_{i=1}^{N}
\log \hat{\mathbf{p}}_i \big[ y_i^{(\ell^\star)} \big].
\label{eq:fuse_loss}
\end{equation}
Since the encoders are aligned across modalities and regularized to respect taxonomy, the fusion predictor is more robust when one modality is degraded.

\paragraph{Degradation Models.}
To assess robustness under realistic perturbations, we introduce degradation strategies for both modalities. For images, we simulate optical blur and defocus by convolving the input $x$ with a normalized averaging kernel $h \in \mathbb{R}^{k \times k}$, where the kernel size $k$ controls the severity of high-frequency attenuation. For DNA, we model sequencing errors and partial reads by transforming a clean sequence $s$ into a corrupted observation $\tilde{s}$ via a stochastic pipeline comprising five operations: (1) \emph{substitution} of nucleotides with probability $p_{\mathrm{sub}}$; (2) \emph{ambiguous masking} where bases are replaced by `N' with probability $p_{\mathrm{mask}}$; (3) \emph{insertions and deletions} ($p_{\mathrm{ins}}, p_{\mathrm{del}}$) to simulate frameshifts; (4) \emph{contiguous dropout} of a subsequence with relative length $\rho$; and (5) \emph{tail truncation} ($\tau$) to mimic incomplete reads.

\begin{algorithm}[t]
\caption{CLiBD-HiR: End-to-End Hierarchy-Guided Multimodal Contrastive Training}
\label{alg:clibd_hir_e2e}
\begin{algorithmic}[1]
\Require Encoders $f_V,f_D,f_T$; temperature $\tau$; weights $\lambda_{VT},\lambda_{VD},\lambda_{DT},\lambda_{\text{HiR}}$; hierarchy weights $\{\alpha_\ell\}_{\ell=1}^{L}$
\While{not converged}
    \State Sample mini-batch $\{(x_i^{V},x_i^{D},x_i^{T},\mathbf{y}_i)\}_{i=1}^{N}$
    \For{$i = 1$ to $N$}
        \State $\mathbf{v}_i \gets f_V(x_i^{V})$, \ $\mathbf{d}_i \gets f_D(x_i^{D})$, \ $\mathbf{t}_i \gets f_T(x_i^{T})$
        \State (Optional) $\ell_2$-normalize $\mathbf{v}_i,\mathbf{d}_i,\mathbf{t}_i$
    \EndFor

    \State Compute pairwise similarities with temperature $\tau$
    \State Compute symmetric CLIP-style losses $\mathcal{L}_{V\leftrightarrow T}$, $\mathcal{L}_{V\leftrightarrow D}$, $\mathcal{L}_{D\leftrightarrow T}$
    \State $\mathcal{L}_{\text{CLIBD}} \gets \lambda_{VT}\mathcal{L}_{V\leftrightarrow T} + \lambda_{VD}\mathcal{L}_{V\leftrightarrow D} + \lambda_{DT}\mathcal{L}_{D\leftrightarrow T}$

    \State $\mathcal{L}_{\text{HiR}} \gets 0$, \ $m^{(0)} \gets 0$
    \For{$\ell = 1$ to $L$} \Comment{taxonomy levels: order, family, genus, species}
        \For{$i = 1$ to $N$}
            \State $\mathcal{P}_i^{(\ell)} \gets \{j\neq i \mid y_j^{(\ell)}=y_i^{(\ell)}\}$,\ \ 
                   $\mathcal{N}_i^{(\ell)} \gets \{k\neq i \mid y_k^{(\ell)}\neq y_i^{(\ell)}\}$
        \EndFor
        \State Compute $\ell^{(\ell)}(i,j)$ using Eq.~\ref{eq:pair_loss_level}
        \State $m^{(\ell)} \gets \max_{i,j \in \mathcal{P}_i^{(\ell)}} \ell^{(\ell)}(i,j)$
        \If{$\ell=1$}
            \State $\tilde{\ell}^{(\ell)}(i,j) \gets \ell^{(\ell)}(i,j)$
        \Else
            \State $\tilde{\ell}^{(\ell)}(i,j) \gets \max\!\big(\ell^{(\ell)}(i,j),\, m^{(\ell-1)}\big)$
        \EndIf
        \State $\mathcal{L}_{\text{HiR}}^{(\ell)} \gets \frac{1}{N}\sum_{i=1}^{N}\frac{1}{|\mathcal{P}_i^{(\ell)}|}\sum_{j\in \mathcal{P}_i^{(\ell)}} \tilde{\ell}^{(\ell)}(i,j)$
        \State $\mathcal{L}_{\text{HiR}} \gets \mathcal{L}_{\text{HiR}} + \alpha_\ell \mathcal{L}_{\text{HiR}}^{(\ell)}$
    \EndFor

    \State $\mathcal{L}_{\text{total}} \gets \mathcal{L}_{\text{CLIBD}} + \lambda_{\text{HiR}}\mathcal{L}_{\text{HiR}}$
    \State Update parameters of $f_V,f_D,f_T$ using $\nabla \mathcal{L}_{\text{total}}$
\EndWhile
\State \Return Trained encoders $f_V,f_D,f_T$
\end{algorithmic}
\end{algorithm}

\begin{algorithm}[t]
\caption{CLiBD-HiR-Fuse: End-to-End Multimodal Contrastive + HiR + Fusion-Supervised Training}
\label{alg:clibd_hir_fuse_e2e}
\begin{algorithmic}[1]
\Require Encoders $f_V,f_D,f_T$; fusion head $g_\phi$; temperature $\tau$;
weights $\lambda_{VT},\lambda_{VD},\lambda_{DT},\lambda_{\text{HiR}},\lambda_{\text{fuse}}$;
hierarchy weights $\{\alpha_\ell\}_{\ell=1}^{L}$
\While{not converged}
    \State Sample mini-batch $\{(x_i^{V},x_i^{D},x_i^{T},\mathbf{y}_i)\}_{i=1}^{N}$
    \For{$i = 1$ to $N$}
        \State $\mathbf{v}_i \gets f_V(x_i^{V})$, \ $\mathbf{d}_i \gets f_D(x_i^{D})$, \ $\mathbf{t}_i \gets f_T(x_i^{T})$
        \State (Optional) $\ell_2$-normalize $\mathbf{v}_i,\mathbf{d}_i,\mathbf{t}_i$
        \State $\mathbf{z}_i \gets [\mathbf{v}_i\,;\,\mathbf{d}_i]$ \Comment{fusion input}
        \State $\hat{\mathbf{p}}_i \gets g_\phi(\mathbf{z}_i)$ \Comment{taxonomy logits/probabilities}
    \EndFor

    \State Compute similarities with temperature $\tau$
    \State Compute symmetric CLIP-style losses $\mathcal{L}_{V\leftrightarrow T}$, $\mathcal{L}_{V\leftrightarrow D}$, $\mathcal{L}_{D\leftrightarrow T}$
    \State $\mathcal{L}_{\text{CLIBD}} \gets \lambda_{VT}\mathcal{L}_{V\leftrightarrow T} + \lambda_{VD}\mathcal{L}_{V\leftrightarrow D} + \lambda_{DT}\mathcal{L}_{D\leftrightarrow T}$

    \State Compute hierarchy-guided loss $\mathcal{L}_{\text{HiR}}$ exactly as in Alg.~\ref{alg:clibd_hir_e2e}
    \State Compute fusion classification loss $\mathcal{L}_{\text{fuse}}$ (Eq.~\ref{eq:fuse_loss})

    \State $\mathcal{L}_{\text{total}} \gets 
           \mathcal{L}_{\text{CLIBD}} + \lambda_{\text{HiR}}\mathcal{L}_{\text{HiR}} + \lambda_{\text{fuse}}\mathcal{L}_{\text{fuse}}$
    \State Update parameters of $f_V,f_D,f_T,g_\phi$ using $\nabla \mathcal{L}_{\text{total}}$
\EndWhile
\State \Return Trained encoders $f_V,f_D,f_T$ and fusion head $g_\phi$
\end{algorithmic}
\end{algorithm}

\section{Experiments}

\paragraph{Dataset and split.}
We use the BIOSCAN-1M \cite{bioscan1m} insect dataset and construct paired samples consisting of a specimen image, a COI DNA barcode sequence, and a textual taxonomy description derived from the hierarchical labels (order, family, genus, species). Each sample includes both string taxonomy fields and consistent integer label IDs at each taxonomic level. Our final split contains 903,536 training samples and 224,777 test samples. The split is \emph{closed-set} across all taxonomic levels: every order, family, genus, and species that appears in the test set also appears in the training set (no unseen taxa in test). Taxonomic completeness varies across specimens (many are labeled only up to coarser levels such as order or family); accordingly, we report evaluation at each level using all available labels for that level.

\subsection{Baselines and evaluation metrics}

We use CLIBD \cite{gong2024clibd} as the primary reference and construct all comparisons to isolate the effects of hierarchical regularization and fusion.
\textbf{(1) No-fusion comparison (Table~\ref{tab:nofusion_clibd_vs_hir}).} We report \textbf{CLIBD} and \textbf{CLIBD-HiR} under clean and noisy settings, evaluating Image$\rightarrow$Text and DNA$\rightarrow$Text Top-1/Top-5 taxonomic classification accuracy without any Image--DNA fusion.
\textbf{(2) Fusion comparison within CLIBD-HiR (Table~\ref{tab:taxa_classification}).} Here we keep the same CLIBD-HiR training setup and compare three variants: \textbf{CLIBD-HiR} (no fusion; unimodal I$\rightarrow$T and D$\rightarrow$T), \textbf{CLIBD-HiR + Avg} (na\"ive averaging of image and DNA embeddings for I+D$\rightarrow$T), and \textbf{CLIBD-HiR + Fusion} (our learned GatedFusion head for I+D$\rightarrow$T). We evaluate robustness under both DNA-only noise (\emph{Noisy D}) and joint image+DNA noise (\emph{Noisy I+D}). This isolates the benefit of adaptive fusion beyond hierarchy-aware representation learning. As per the metrics, we report Top-1 and Top-5 taxonomic prediction accuracy at four hierarchical levels (order, family, genus, species). For each query (image, DNA, or fused image+DNA), we rank candidate taxonomy text prompts by similarity in the shared embedding space and measure whether the ground-truth label appears at rank 1 or within the top 5. \textbf{Global} denotes an aggregate accuracy across the four taxonomic levels.

\paragraph{Models and training setup.}
We adopt a three-encoder architecture consisting of an image encoder, a DNA encoder, and a text encoder.
The image and text encoders are initialized from a pretrained vision--language model, either standard OpenCLIP ViT-L/14~\citep{cherti2023openclip_scaling} or BioCLIP ViT-L/14~\citep{stevens2024bioclip}. The DNA branch is built upon DNABERT2~\citep{zhou2024dnabert2}.
When BioCLIP is used, we employ a modified DNABERT2 variant by adding a learnable linear projection layer on top of the DNABERT2 embedding
to match the embedding dimension of the image--text backbone. Following Algo.~1 and Algo.~2, we train the image and DNA encoders end-to-end. The optimization strategy for the text encoder depends on the chosen backbone. With BioCLIP, we freeze the entire text encoder. With standard OpenCLIP, we fine-tune the text encoder during training. Empirically, freezing BioCLIP text encoder improves performance, likely because BioCLIP already encodes strong biological language priors \citep{stevens2024bioclip}; this differs from the original CLIBD training recipe, which fine-tunes the text encoder \citep{gong2024clibd}. Unless stated otherwise, we report results using this \emph{fixed-BioCLIP-text} variant. Training uses contrastive alignment losses together with a hierarchy-aware loss on augmented image features \citep{khosla2020supcon,zhang2022usealllabels}. For the fusion variant, we add a trainable \textit{GatedFusion} head, a lightweight 2-layer MLP (Linear--ReLU--Dropout--Linear--Sigmoid) that predicts a per-dimension gate from concatenated image and DNA embeddings and mixes them before normalization, supervised by an additional fused-to-text contrastive loss.

\paragraph{Implementation and noise settings.}
We train with distributed data parallel on 4 NVIDIA A100 GPUs for $\sim$1 day (batch size 30, 10 epochs). Images are loaded from an HDF5 container and paired with DNA barcodes and taxonomy text from CSV metadata. Optimization uses AdamW with a OneCycle schedule (base LR $1{\times}10^{-6}$, max LR $5{\times}10^{-5}$), with $\lambda_{\text{HiR}}=0.99$ and $\alpha=0.1$ in the gathered contrastive loss; the fusion variant adds a fused-to-text term weighted by $\lambda_{\text{fuse}}=0.7$ (omitted for the no-fusion baseline). For robustness evaluation, we apply \emph{inference-time} modality degradation only (no noisy data during training): DNA is corrupted with substitutions ($p_{\text{sub}}{=}0.01$), insertions/deletions ($p_{\text{ins}}{=}p_{\text{del}}{=}0.002$), masking to \texttt{N} ($p_{\text{mask}}{=}0.003$), contiguous \texttt{N}-dropout (run fraction $0.05$), and tail truncation (10\%), while images use blur noise with a $7{\times}7$ kernel. We report clean, DNA-noisy, and joint image+DNA noisy results.

\begin{table*}[t]
\centering
\caption{\textbf{Algo~1 (CLIBD-HiR): no-fusion evaluation.} Comparison of CLIBD and CLIBD-HiR without any image--DNA fusion module. We report \textbf{Top-1 / Top-5} taxonomic prediction accuracy (\%) for Image$\!\rightarrow$Text and DNA$\!\rightarrow$Text under \textbf{clean} inputs and under \textbf{noisy} inputs (synthetically degraded at inference). \textbf{Global} denotes an aggregate across order, family, genus, and species. \colorbox{lightgray}{Highlighted rows} indicate our proposed HiR model.}
\label{tab:nofusion_clibd_vs_hir}

\footnotesize
\renewcommand{\arraystretch}{1.15}
\setlength{\tabcolsep}{4pt} 

\begin{tabular*}{\textwidth}{@{\extracolsep{\fill}}ll l ccccc}
\toprule
\multirow{2}{*}{\textbf{Setting}} & \multirow{2}{*}{\textbf{Method}} & \multirow{2}{*}{\textbf{Cond.}} 
& \multicolumn{5}{c}{\textbf{Top-1 / Top-5 Accuracy}} \\
\cmidrule(l){4-8}
& & & \textbf{Order} & \textbf{Family} & \textbf{Genus} & \textbf{Species} & \textbf{Global} \\
\midrule

\multirow{2}{*}{I$\!\rightarrow\!$T} & CLIBD & Clean & 98.8 / 99.2 & 93.3 / 95.8 & \textbf{77.8} / \textbf{90.4} & \textbf{46.6} / \textbf{71.4} & 75.5 / \textbf{89.8} \\
& & Noisy & \textbf{88.8} / \textbf{92.1} & 66.8 / 74.7 & \textbf{48.7} / \textbf{65.6} & \textbf{22.2} / \textbf{42.0} & 40.0 / 58.4 \\
\cmidrule(l){2-8}

\rowcolor{lightgray}
& \textbf{CLIBD-HiR} & Clean & \textbf{99.3} / \textbf{99.5} & \textbf{94.3} / \textbf{96.1} & 76.8 / 89.1 & 46.6 / 68.3 & \textbf{78.2} / 89.2 \\
\rowcolor{lightgray}
\multirow{-2}{*}{I$\!\rightarrow\!$T} & \textbf{CLIBD-HiR} & Noisy & 88.7 / 91.4 & \textbf{70.1} / \textbf{77.0} & 42.9 / 59.0 & 15.5 / 31.8 & \textbf{46.6} / \textbf{59.6} \\

\midrule

\multirow{2}{*}{D$\!\rightarrow\!$T} & CLIBD & Clean & 99.9 / 99.9 & \textbf{99.4} / 99.8 & 94.7 / 98.1 & 68.1 / 88.5 & 94.8 / 98.3 \\
& & Noisy & 99.7 / \textbf{99.9} & 57.1 / 97.8 & \textbf{81.6} / 94.2 & 48.6 / 76.3 & 52.4 / 91.5 \\
\cmidrule(l){2-8}

\rowcolor{lightgray}
& \textbf{CLIBD-HiR} & Clean & \textbf{100.0} / \textbf{100} & 99.3 / \textbf{99.9} & \textbf{95.9} / \textbf{98.8} & \textbf{71.6} / \textbf{91.3} & \textbf{95.6} / \textbf{98.7} \\
\rowcolor{lightgray}
\multirow{-2}{*}{D$\!\rightarrow\!$T} & \textbf{CLIBD-HiR} & Noisy & \textbf{99.8} / \textbf{99.9} & \textbf{70.3} / \textbf{99.2} & 81.5 / \textbf{95.1} & \textbf{51.6} / \textbf{79.3} & \textbf{66.0} / \textbf{96.9} \\

\bottomrule
\end{tabular*}
\end{table*}

\paragraph{Results.}
Table~\ref{tab:nofusion_clibd_vs_hir} shows that adding HiR to CLIBD improves no-fusion taxonomic prediction, with the largest gains under noise. For I$\!\rightarrow\!$T, CLIBD-HiR increases Global Top-1 from 75.5 to 78.2 on clean data and from 40.0 to 46.6 on noisy data (Top-5: 58.4 to 59.6), driven mainly by improved coarse-level accuracy (e.g., noisy Family Top-1: 66.8 to 70.1). For D$\!\rightarrow\!$T, HiR yields a small clean gain (Global Top-1: 94.8 to 95.6) but a substantial robustness improvement under noisy DNA (Global Top-1: 52.4 to 66.0; Global Top-5: 91.5 to 96.9), with a notable increase at the family level (57.1 to 70.3 Top-1). Overall, HiR consistently improves global performance and noise robustness without using fusion. Moreover, Table~\ref{tab:taxa_classification} analyzes fusion within the same CLIBD-HiR training setup. The unimodal baselines (I$\!\rightarrow\!$T and D$\!\rightarrow\!$T) establish modality-specific performance, while the third block, \textbf{I+D$\!\rightarrow\!$T (Avg)}, fuses image and DNA by simple embedding averaging (no fusion module). Our learned fusion model, \textbf{I+D$\!\rightarrow\!$T (Ours)}, improves over averaging in the most realistic setting where both modalities are noisy: Global accuracy increases from 85.5/96.5 to 88.0/97.5 (Top-1/Top-5), with the largest gain at species level (54.6/79.9 to 57.4/81.7). Under DNA-only noise, averaging and learned fusion are comparable globally (91.3/97.7 vs.\ 91.4/98.0), indicating that the main benefit of the fusion module is robustness when image and DNA quality vary simultaneously.

\begin{table*}[t]
\centering
\caption{\textbf{Algo~2 (CLIBD-HiR-Fuse): fusion evaluation.} We report \textbf{Top-1 / Top-5} taxonomic prediction accuracy (\%). \colorbox{lightgray}{Highlighted rows} indicate our proposed fusion model. \textbf{Bold} indicates the best performance for that specific condition (Clean, Noisy D, or Noisy I+D) across all methods.}
\label{tab:taxa_classification}

\footnotesize
\renewcommand{\arraystretch}{1.15} 
\setlength{\tabcolsep}{3pt} 

\begin{tabular*}{\textwidth}{@{\extracolsep{\fill}}ll ccccc}
\toprule
\multirow{2}{*}{\textbf{Method}} & \multirow{2}{*}{\textbf{Cond.}} 
& \multicolumn{5}{c}{\textbf{Top-1 / Top-5 Accuracy}} \\
\cmidrule(l){3-7}
& & \textbf{Order} & \textbf{Family} & \textbf{Genus} & \textbf{Species} & \textbf{Global} \\
\midrule

\multirow{2}{*}{I$\!\rightarrow\!$T (Ours)}
& Clean & 99.3 / 99.5 & 93.5 / 99.9 & 74.7 / 98.9 & 40.4 / 91.8 & 77.0 / 98.7 \\
& Noisy I & 88.3 / 91.8 & 67.0 / 75.9 & 37.0 / 55.7 & 11.7 / 26.7 & 50.2 / 62.1 \\
\cmidrule(l){1-7}

\multirow{2}{*}{D$\!\rightarrow\!$T (Ours)}
& Clean & \textbf{100.0} / \textbf{100} & 99.5 / \textbf{99.9} & \textbf{96.1} / \textbf{98.9} & \textbf{74.4} / \textbf{92.7} & 95.7 / 98.7 \\
& Noisy D & 99.8 / 99.9 & 87.3 / 99.3 & 86.1 / 95.6 & 57.8 / 82.0 & 82.3 / 97.1 \\
\cmidrule(l){1-7}

\multirow{3}{*}{I+D$\!\rightarrow\!$T (Avg)}
& Clean & 99.9 / \textbf{100} & 99.6 / \textbf{99.9} & 95.5 / 98.6 & 71.2 / 89.8 & 95.7 / 98.6 \\
& Noisy D & 99.9 / \textbf{100} & \textbf{96.8} / \textbf{99.7} & \textbf{90.6} / \textbf{97.1} & 60.1 / 82.9 & 91.3 / 97.7 \\
& Noisy I+D & 99.6 / \textbf{100} & 91.3 / 99.3 & \textbf{87.9} / \textbf{96.1} & 54.6 / 79.9 & 85.5 / 96.5 \\
\cmidrule(l){1-7}

\rowcolor{lightgray}
\textbf{I+D$\!\rightarrow\!$T (Ours)}
& Clean & \textbf{100.0} / \textbf{100} & \textbf{99.7} / \textbf{99.9} & 96.0 / \textbf{98.9} & 73.5 / 91.2 & \textbf{96.1} / \textbf{98.8} \\

\rowcolor{lightgray}
\textbf{(GatedFusion)}
& Noisy D & \textbf{100.0} / \textbf{100} & 96.2 / \textbf{99.7} & 88.7 / 96.9 & \textbf{60.8} / \textbf{83.2} & \textbf{91.4} / \textbf{98.0} \\

\rowcolor{lightgray}
& Noisy I+D & \textbf{99.9} / \textbf{100} & \textbf{93.0} / \textbf{99.5} & 87.0 / 96.0 & \textbf{57.4} / \textbf{81.7} & \textbf{88.0} / \textbf{97.5} \\

\bottomrule
\end{tabular*}
\end{table*}

\paragraph{Limitations.}
We do not use the original CLIBD split, which is designed around seen/unseen evaluation across multiple taxonomic levels and is highly imbalanced. In our initial experiments, this split substantially reduces the effective supervision available at deeper levels (genus/species) and makes hierarchy learning less informative, leading to performance that is close to the baseline. To clearly demonstrate the impact of hierarchy-aware regularization and fusion, we therefore report results on our split where all taxonomic levels are consistently represented. Developing training and evaluation protocols that better handle the severe long-tail imbalance and level-dependent seen/unseen structure in CLIBD remains an important direction for future work.

\section{Conclusion}
We presented HiR-Fusion, a hierarchy-guided multimodal framework for robust taxonomic prediction from images, DNA barcodes, and their combination. Our first variant (CLIBD-HiR) improves noise robustness by explicitly shaping embedding geometry to respect biological hierarchy, and our second variant (CLIBD-HiR-Fuse) further enhances performance by learning an adaptive fusion module that outperforms naive averaging, particularly when both modalities are degraded. Across clean and noisy settings, our results show that incorporating taxonomic structure and reliability-aware fusion yields more robust and practically useful biodiversity recognition models.

\bibliographystyle{iclr2026_conference}
\bibliography{iclr2026_conference}

@inproceedings{bioscan1m,
  title     = {A Step Towards Worldwide Biodiversity Assessment: The {BIOSCAN-1M} Insect Dataset},
  booktitle = {Advances in Neural Information Processing Systems},
  author    = {Gharaee, Zahra and Gong, ZeMing and Pellegrino, Nicholas and Zarubiieva, Iuliia and
               Haurum, Joakim Bruslund and Lowe, Scott C. and McKeown, Jaclyn T. A. and
               Ho, Chris C. Y. and McLeod, Joschka and Wei, Yi-Yun C. and Agda, Jireh and
               Ratnasingham, Sujeevan and Steinke, Dirk and Chang, Angel X. and Taylor, Graham W. and Fieguth, Paul},
  year      = {2023},
  volume    = {36},
  pages     = {43593--43619},
  publisher = {Curran Associates, Inc.}
}

@article{gong2024clibd,
  author        = {Gong, ZeMing and Wang, Austin T. and Huo, Xiaoliang and Haurum, Joakim Bruslund and
                   Lowe, Scott C. and Taylor, Graham W. and Chang, Angel X.},
  title         = {{CLIBD}: Bridging Vision and Genomics for Biodiversity Monitoring at Scale},
  journal       = {arXiv preprint},
  year          = {2024},
  eprint        = {2405.17537},
  archivePrefix = {arXiv},
  primaryClass  = {cs.AI},
  doi           = {10.48550/arXiv.2405.17537}
}

@inproceedings{stevens2024bioclip,
  title     = {{B}io{CLIP}: A Vision Foundation Model for the Tree of Life},
  author    = {Stevens, Samuel and Wu, Jiaman and Thompson, Matthew J. and Campolongo, Elizabeth G. and
               Song, Chan Hee and Carlyn, David Edward and Dong, Li and Dahdul, Wasila M. and
               Stewart, Charles and Berger-Wolf, Tanya and Chao, Wei-Lun and Su, Yu},
  booktitle = {Proceedings of the IEEE/CVF Conference on Computer Vision and Pattern Recognition (CVPR)},
  year      = {2024}
}

@inproceedings{radford2021clip,
  title     = {Learning Transferable Visual Models From Natural Language Supervision},
  author    = {Radford, Alec and Kim, Jong Wook and Hallacy, Chris and Ramesh, Aditya and
               Goh, Gabriel and Agarwal, Sandhini and Sastry, Girish and Askell, Amanda and
               Mishkin, Pamela and Clark, Jack and Krueger, Gretchen and Sutskever, Ilya},
  booktitle = {Proceedings of the 38th International Conference on Machine Learning (ICML)},
  series    = {Proceedings of Machine Learning Research},
  volume    = {139},
  pages     = {8748--8763},
  year      = {2021},
  publisher = {PMLR}
}

@inproceedings{cherti2023openclip_scaling,
  title     = {Reproducible Scaling Laws for Contrastive Language-Image Learning},
  author    = {Cherti, Mehdi and Beaumont, Romain and Wightman, Ross and Wortsman, Mitchell and
               Ilharco, Gabriel and Gordon, Cade and Schuhmann, Christoph and Schmidt, Ludwig and Jitsev, Jenia},
  booktitle = {Proceedings of the IEEE/CVF Conference on Computer Vision and Pattern Recognition (CVPR)},
  year      = {2023},
  pages     = {2818--2829},
  doi       = {10.1109/CVPR52729.2023.00276},
  eprint    = {2212.07143},
  archivePrefix = {arXiv},
  primaryClass  = {cs.CV}
}

@inproceedings{zhou2024dnabert2,
  title     = {{DNABERT-2}: Efficient Foundation Model and Benchmark for Multi-Species Genomes},
  author    = {Zhou, Zhihan and Ji, Yanrong and Li, Weijian and Dutta, Pratik and Davuluri, Ramana V. and Liu, Han},
  booktitle = {International Conference on Learning Representations (ICLR)},
  year      = {2024},
  eprint    = {2306.15006},
  archivePrefix = {arXiv},
  primaryClass  = {cs.LG},
  doi       = {10.48550/arXiv.2306.15006}
}

@inproceedings{zhang2022usealllabels,
  title     = {Use All The Labels: A Hierarchical Multi-Label Contrastive Learning Framework},
  author    = {Zhang, Shu and Xu, Ran and Xiong, Caiming and Ramaiah, Chetan},
  booktitle = {Proceedings of the IEEE/CVF Conference on Computer Vision and Pattern Recognition (CVPR)},
  year      = {2022},
  eprint    = {2204.13207},
  archivePrefix = {arXiv},
  primaryClass  = {cs.CV},
  doi       = {10.48550/arXiv.2204.13207}
}

@inproceedings{khosla2020supcon,
  title     = {Supervised Contrastive Learning},
  author    = {Khosla, Prannay and Teterwak, Piotr and Wang, Chen and Sarna, Aaron and Tian, Yonglong and
               Isola, Phillip and Maschinot, Aaron and Liu, Ce and Krishnan, Dilip},
  booktitle = {Advances in Neural Information Processing Systems},
  year      = {2020},
  eprint    = {2004.11362},
  archivePrefix = {arXiv},
  primaryClass  = {cs.LG}
}

@article{ratnasingham2007bold,
  title={BOLD: The Barcode of Life Data System (http://www. barcodinglife. org)},
  author={Ratnasingham, Sujeevan and Hebert, Paul DN},
  journal={Molecular ecology notes},
  volume={7},
  number={3},
  pages={355--364},
  year={2007},
  publisher={Wiley Online Library}
}

@article{meiklejohn2019assessment,
  title={Assessment of BOLD and GenBank--Their accuracy and reliability for the identification of biological materials},
  author={Meiklejohn, Kelly A and Damaso, Natalie and Robertson, James M},
  journal={PloS one},
  volume={14},
  number={6},
  pages={e0217084},
  year={2019},
  publisher={Public Library of Science San Francisco, CA USA}
}

@article{medigue1999detecting,
  title={Detecting and analyzing DNA sequencing errors: toward a higher quality of the Bacillus subtilis genome sequence},
  author={M{\'e}digue, Claudine and Rose, Matthias and Viari, Alain and Danchin, Antoine},
  journal={Genome research},
  volume={9},
  number={11},
  pages={1116--1127},
  year={1999},
  publisher={Cold Spring Harbor Lab}
}

@article{nguyen2024sawit,
  title={SAWIT: A small-sized animal wild image dataset with annotations},
  author={Nguyen, Thi Thu Thuy and Eichholtzer, Anne C and Driscoll, Don A and Semianiw, Nathan I and Corva, Dean M and Kouzani, Abbas Z and Nguyen, Thanh Thi and Nguyen, Duc Thanh},
  journal={Multimedia Tools and Applications},
  volume={83},
  number={11},
  pages={34083--34108},
  year={2024},
  publisher={Springer}
}

@inproceedings{van2018inaturalist,
  title={The inaturalist species classification and detection dataset},
  author={Van Horn, Grant and Mac Aodha, Oisin and Song, Yang and Cui, Yin and Sun, Chen and Shepard, Alex and Adam, Hartwig and Perona, Pietro and Belongie, Serge},
  booktitle={Proceedings of the IEEE conference on computer vision and pattern recognition},
  pages={8769--8778},
  year={2018}
}

@article{cong2017coi,
  title={When COI barcodes deceive: complete genomes reveal introgression in hairstreaks},
  author={Cong, Qian and Shen, Jinhui and Borek, Dominika and Robbins, Robert K and Opler, Paul A and Otwinowski, Zbyszek and Grishin, Nick V},
  journal={Proceedings of the Royal Society B: Biological Sciences},
  volume={284},
  number={1848},
  pages={20161735},
  year={2017},
  publisher={The Royal Society}
}

@inproceedings{girdhar2023imagebind,
  title={Imagebind: One embedding space to bind them all},
  author={Girdhar, Rohit and El-Nouby, Alaaeldin and Liu, Zhuang and Singh, Mannat and Alwala, Kalyan Vasudev and Joulin, Armand and Misra, Ishan},
  booktitle={Proceedings of the IEEE/CVF conference on computer vision and pattern recognition},
  pages={15180--15190},
  year={2023}
}

@InProceedings{Gao_2024_CVPR,
  author    = {Gao, Zixian and Jiang, Xun and Xu, Xing and Shen, Fumin and Li, Yujie and Shen, Heng Tao},
  title     = {Embracing Unimodal Aleatoric Uncertainty for Robust Multimodal Fusion},
  booktitle = {Proceedings of the IEEE/CVF Conference on Computer Vision and Pattern Recognition (CVPR)},
  year      = {2024},
  pages     = {26876--26885}
}

@inproceedings{Cao_2023_ICCV,
  author    = {Cao, Bing and Sun, Yiming and Zhu, Pengfei and Hu, Qinghua},
  title     = {Multi-Modal Gated Mixture of Local-to-Global Experts for Dynamic Image Fusion},
  booktitle = {Proceedings of the IEEE/CVF International Conference on Computer Vision (ICCV)},
  year      = {2023},
  pages     = {23555--23564}
}

@inproceedings{Han_2024_FuseMoE,
  author    = {Han, Xing and Nguyen, Huy and Harris, Carl and Ho, Nhat and Saria, Suchi},
  title     = {FuseMoE: Mixture-of-Experts Transformers for Fleximodal Fusion},
  booktitle = {Advances in Neural Information Processing Systems (NeurIPS)},
  year      = {2024}
}

\end{document}